\documentclass[conference]{IEEEtran}
\IEEEoverridecommandlockouts

\usepackage{cite}
\usepackage{amsmath,amssymb,amsfonts}
\usepackage{algorithmic}
\usepackage{graphicx}
\usepackage{textcomp}
\usepackage{xcolor}
\usepackage{caption}
\usepackage{subcaption}
\usepackage{hyperref}
\usepackage{epstopdf}

\def\BibTeX{{\rm B\kern-.05em{\sc i\kern-.025em b}\kern-.08em
    T\kern-.1667em\lower.7ex\hbox{E}\kern-.125emX}}
\begin{document}

%%%%%%%%%%%%%%%%%%%%%%%%%%%%%%%%%%%%%%%%%%%%%%%%%%%%%%%%%%%%%%%%%%%%%%%%%%%%%%%%%%%%%%
%%%%%%%%%%%%%%%%%%%%%%%%%%%%%%%%%%%%%%%%%%%%%%%%%%%%%%%%%%%%%%%%%%%%%%%%%%%%%%%%%%%%%%
%%%%%%%%%%%%%%%%%%%%%%%%%%%%% Cover Page information %%%%%%%%%%%%%%%%%%%%%%%%%%%%%%%%%
%%%%%%%%%%%%%%%%%%%%%%%%%%%%%%%%%%%%%%%%%%%%%%%%%%%%%%%%%%%%%%%%%%%%%%%%%%%%%%%%%%%%%%
%%%%%%%%%%%%%%%%%%%%%%%%%%%%%%%%%%%%%%%%%%%%%%%%%%%%%%%%%%%%%%%%%%%%%%%%%%%%%%%%%%%%%%

%%%%%% Title %%%%%%%%%%
\title{Hierarchical confusion matrix for classification performance evaluation}

%%%%%% Authors %%%%%%%%
\author{
    \IEEEauthorblockN{1\textsuperscript{st} Kevin Riehl}
    \IEEEauthorblockA{\textit{Gurdon Institute, University of Cambridge, UK} \\
    \textit{Technische Universität Darmstadt, Germany} \\
    \textit{kevin.riehl@gurdon.cam.ac.uk}}
    \and
    
    \IEEEauthorblockN{2\textsuperscript{nd} Michael Neunteufel}
    \IEEEauthorblockA{\textit{Institute of Analysis and Scientific Computing} \\
    \textit{Technische Universität Wien} \\
    \textit{Vienna, Austria} \\
    \textit{michael.neunteufel@tuwien.ac.at} }
    \and
    
    \IEEEauthorblockN{3\textsuperscript{rd} Martin Hemberg}
    \IEEEauthorblockA{\textit{Evergrande Center for Immunological Disease} \\
    \textit{Brigham and Women’s Hospital, Harvard Medical School} \\
    \textit{75 Francis Street, Boston, MA 02215, USA} \\
    \textit{mhemberg@bwh.harvard.edu} }    
}

\maketitle

%%%%%% Abstract %%%%%%%%%%
\begin{abstract}
In this work we propose a novel concept of a hierarchical confusion matrix, opening the door for popular confusion matrix based (flat) evaluation measures from binary classification problems, while considering the peculiarities of hierarchical classification problems.
We develop the concept to a generalized form and prove its applicability to all types of hierarchical classification problems including directed acyclic graphs, multi path labelling, and non mandatory leaf node prediction.
Finally, we use measures based on the novel confusion matrix to evaluate models within a benchmark for three real world hierarchical classification applications and compare the results to established evaluation measures. The results outline the reasonability of this approach and its usefulness to evaluate hierarchical classification problems.
The implementation of hierarchical confusion matrix is available on GitHub.
\end{abstract}

%%%%% Keywords %%%%%%%%%%

%%%%%%%%%%%%%%%%%%%%%%%%%%%%%%%%%%%%%%%%%%%%%%%%%%%%%%%%%%%%%%%%%%%%%%%%%%%%%%%%%%%%%%
%%%%%%%%%%%%%%%%%%%%%%%%%%%%%%%%%%%%%%%%%%%%%%%%%%%%%%%%%%%%%%%%%%%%%%%%%%%%%%%%%%%%%%
%%%%%%%%%%%%%%%%%%%%%%%%%%%%% Introduction %%%%%%%%%%%%%%%%%%%%%%%%%%%%%%%%%%%%%%%%%%%
%%%%%%%%%%%%%%%%%%%%%%%%%%%%%%%%%%%%%%%%%%%%%%%%%%%%%%%%%%%%%%%%%%%%%%%%%%%%%%%%%%%%%%
%%%%%%%%%%%%%%%%%%%%%%%%%%%%%%%%%%%%%%%%%%%%%%%%%%%%%%%%%%%%%%%%%%%%%%%%%%%%%%%%%%%%%%
\section{Introduction}

% Definition of hierarchical classification
Hierarchical classification is the task of assigning $m$ multiple classes out of $n$ multiple, overlapping classes ($C_i$) disposed in a hierarchical structure. Classes in the hierarchical classification task can be divided into sub classes or grouped into super classes.
Often, hierarchical classification is wrongly referred to other problems\cite{silla2011survey, kiritchenko2006learning} such as hierarchical clustering of objects to generate structures of classes~\cite{gordon1987review,gauch1981hierarchical}. This work explicitly understands hierarchical classification as allocation of objects to a set of given classes rather than the design of classes and their structure itself. 

%What are remaining questions in hierarchical classification research on evaluation?
% Überschwenk zu Evaluation measures
Machine learning research started to focus on (hierarchically) structured classification problems in the context of text categorization in the mid-1990s~\cite{koller1997hierarchically}. From then on, hierarchical classification has been applied in many fields, namely text categorization, music genre classification and bioinformatic applications including sequence classification, and protein function prediction~\cite{silla2011survey,freitas2007tutorial}.
Despite of intensive research, many related questions in the field remain open, as stated e.g. in~\cite{kosmopoulos2015evaluation}.

Hierarchical classification evaluation is part of ongoing research and various measures have been proposed so far ~\cite{costa2007review,sokolova2006beyond}. While established evaluation measures for binary and multi-class classification problems exist, there is still the lack of consensus on which measures to employ in the context of hierarchical classification problems~\cite{costa2007review}. 

%What we are doing in this work
Hierarchical classification studies frequently adopted conventional, flat measures from binary classification problems such as accuracy, precision, or recall~\cite{kiritchenko2006learning,costa2007review,kosmopoulos2015evaluation}. Due to their simple structure based on the confusion matrix, these measures are easy accessibly to human inspection. However, flat measures do not appropriately consider the complex peculiarities of hierarchical classification problems. The measures hierarchical precision, recall, and F-score~\cite{kiritchenko2006learning} are observed to frequently occur in recent studies; these measures try to bridge the evaluation of binary and hierarchical classification problems. 

In this work we propose a novel concept for the evaluation of hierarchical classification problems: the hierarchical confusion matrix. 
This approach opens the door for conventional measures in the context of hierarchical classification and to complete the field of hierarchical classification evaluation. 
Further, we extend the concept to a generalized form and prove its applicability to all types of hierarchical classification problems including directed acyclic graphs, multi path label, and non-mandatory leaf-node prediction. 
Finally, we demonstrate the application of measures based on the novel concept for the comparison of classifiers in three exemplary benchmarks related to different types of hierarchical classification problems.

% Commercial for Github
An implementation of the hierarchical confusion matrix in Python programming language, and the examples and experiments of this study are available on GitHub\footnote{\url{https://github.com/DerKevinRiehl/HierarchicalConfusionMatrix}}.

%Brief remainder of this work
The brief remainder of this work is organized as follows. 
Section~\ref{sec:hclassification} formally introduces the problems of hierarchical classification.
In Section~\ref{sec:relatedwork} a summary of proposed evaluation measures for hierarchical classification is presented.
Section~\ref{sec:hConfusion} proposes the novel hierarchical confusion matrix and possible evaluation measures based on it.
In Section~\ref{sec:experiment} the proposed evaluation measure is applied onto experimental data and compared to existing measures.
Finally, Section~\ref{sec:conclusion} concludes this paper and summarizes its contribution.

%%%%%%%%%%%%%%%%%%%%%%%%%%%%%%%%%%%%%%%%%%%%%%%%%%%%%%%%%%%%%%%%%%%%%%%%%%%%%%%%%%%%%%
%%%%%%%%%%%%%%%%%%%%%%%%%%%%%%%%%%%%%%%%%%%%%%%%%%%%%%%%%%%%%%%%%%%%%%%%%%%%%%%%%%%%%%
%%%%%%%%%%%%%%%%%%%%%%%%%%%%% Hierarchical classification %%%%%%%%%%%%%%%%%%%%%%%%%%%%
%%%%%%%%%%%%%%%%%%%%%%%%%%%%%%%%%%%%%%%%%%%%%%%%%%%%%%%%%%%%%%%%%%%%%%%%%%%%%%%%%%%%%%
%%%%%%%%%%%%%%%%%%%%%%%%%%%%%%%%%%%%%%%%%%%%%%%%%%%%%%%%%%%%%%%%%%%%%%%%%%%%%%%%%%%%%%
\section{Hierarchical Classification} \label{sec:hclassification}
\begin{figure}
    \centering
    \includegraphics[width=\linewidth]{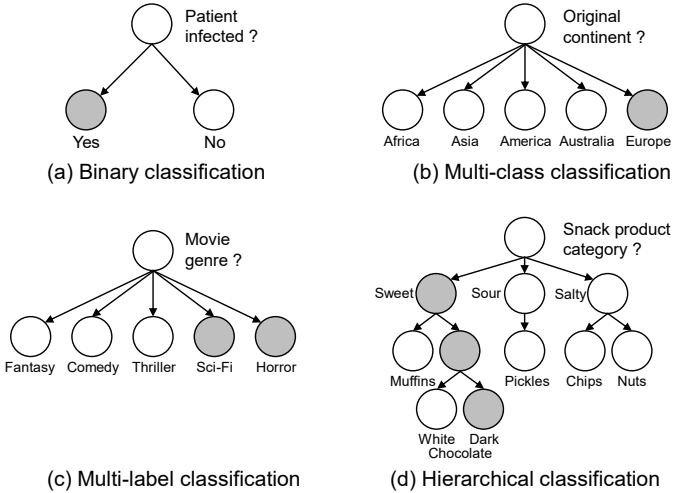}
    \caption{\textbf{Classification problems by example.}    The selected class is marked gray. (a) binary classification. (b) multi-class classification. (c) multi-label classification. (d) hierarchical classification. Note, that the single selected category (\textit{dark chocolate}) causes the selection of multiple parental vertices in the graph on a path up to the root node.}
    \label{fig:classificationProblems}
\end{figure}

% General introduction of different classification tasks
Supervised machine learning allows access to labelled data for training and testing stages.
Classification describes supervised machine learning using categorical labels, and stands for the allocation of objects to a set of existing classes.
The field of classification can be divided into the sub problems binary, multi-class, multi-label, and hierarchical classification, as shown in Fig.~\ref{fig:classificationProblems}. 
In the binary classification problem, objects need to be allocated to one of two non-overlapping classes ($C_1$, $C_2$). 
Multi-class classification requires the choice of one out of $n$ multiple, non-overlapping classes ($C_i$) for a given input object. 
Multi-label classification represents the assignment of objects to $m$ multiple classes out of $n$ multiple, non-overlapping classes ($C_i$).

% Formally define hierarchical classification
Before we define hierarchical classification, let us introduce the notation used in this paper. Therefore, we define a set of data $\Omega$ and a structure $\Gamma$. The structure $\Gamma$ can be described by a graph, where the nodes represent the $n$ classes $C_i \in \Gamma$ and the edges stand for the possible relationships between these classes, namely ancestry and descendance. For each class $C_i$, let us denote the set of all descendants by $D\left \{ C_i \right \}$ and the set of all neighboring nodes (with at least one common ancestor) by $N\left \{ C_i \right \}$. 
%,the set of all ancestors by $A\left \{ C_i \right \}$

The root node $R \in \Gamma$ is the only node without ancestors $A\left \{ R \right \} = \varnothing$. Leaf nodes $i$ are nodes without any descendants.
An allocation path $\vec{P}_i$ is a series of classes within $\Gamma$. This work requires three conditions for $\vec{P}_i$: (1) for each consecutive pair of elements $C_a, C_b \in \vec{P}_i$ the ancestry relationship $C_a \in A\left \{ C_b \right \}$ holds, (2) the connectivity to the root node $R \in \vec{P}_i$, and (3) the connectivity to the allocated class $C_i \in \vec{P}_i$. 
%The span $S\left \{ \vec{P}_i \right \}$ of an allocation path is defined as the number of elements inside the set excluding the root node: $S\left \{ \vec{P}_i \right \} = \left | \vec{P}_i \right | - 1$.
%The structural level $L\left \{ C_i \right \}$ of a class $C_i$ is defined as the span of the allocation path to $C_i$: $L\left \{ C_i \right \} = S\left \{ \vec{P}_i \right \}$.
Let us denote the true class of a given object $o_d \in \Omega$ as $C\left \{ o_d \right \}$.% and its predicted class as $\hat{C}\left \{ o_d \right \}$.

$P_d$ describes the set of all predicted allocation paths for $o_d$, and a single, predicted allocation path is given by $\hat{P}_d \in P_d$. $W_d$ denotes the set of all true allocation paths for $o_d$ based on $\Gamma$ and $\widetilde{P}_d \in W_d$ defines a single, true allocation path. Moreover, let us denote the intersection of $\hat{P}_d$ and $\widetilde{P}_d$ by $\hat{P}_d \cap \widetilde{P}_d$, the (longest, consecutively connected) common path of prediction of $\hat{P}_d$ and $\widetilde{P}_d$ by $\hat{P}_d \: \overline{\cap} \: \widetilde{P}_d$, and the leaf node of a path $\vec{P}_i$ as $Z\left \{ \vec{P}_i \right \}$.

Hierarchical classification is a structured classification problem and represents the allocation of one or multiple paths $\vec{P}_x \in P_d$ within the structure $\Gamma$ to a given object $o_d \in \Omega$. Table~\ref{tab:definitionsA} summarizes the definitions of this work.

\begin{table}
    \centering
    \begin{tabular}{l l}
        \textbf{Symbol} & \textbf{Definition} \\ \hline
        $\Gamma$ & structure of hierarchical classification problem \\
        $\Psi$ & number of allocated paths of classification problem\\
        $\Phi$ & mandatoriness for leaf-node prediction \\ \\
        $\Omega$ & set of data \\
        $o_d$ & a single data set in $O$ \\
        $C_i$ & class $i$, node in $\Gamma$ \\
        $R$ & root node of $\Gamma$ \\
        $C\left \{ o_d \right \}$ & true class of $o_d$ in $\Gamma$ \\
        %$\hat{C}\left \{ o_d \right \}$ & predicted class of $o_d$ in $\Gamma$ \\
        %$A\left \{ C_i \right \}$ & set of ancestors of $C_i$ \\
        $D\left \{ C_i \right \}$ & set of descendants of $C_i$ \\
        $N\left \{ C_i \right \}$ & set of neighbors of $C_i$ \\
        $\vec{P}_i$ & allocation path to $C_i$ \\
        %$S\left \{ \vec{P}_i \right \}$ & span of $\vec{P}_i$ \\
        %$L\left \{ C_i \right \}$ & taxonomic level of $C_i$ \\
        $Z\left \{ \vec{P}_i \right \}$ & leaf node of $\vec{P}_i$ \\
        $P_a \: \overline{\cap} \: P_b$ & longest, consecutively connected, common path\\
        $P_d$ & set of all predicted allocation paths for $o_d$ \\
        $\hat{P}_d$ & a single, predicted allocation path for $o_d$ in $P_d$ \\
        $W_d$ & set of all true allocation paths for $o_d$ \\
        $\widetilde{P}_d$ & a single, true allocation path for $o_d$ in $W_d$ \\
        \\ \hline
    \end{tabular}
    \caption{Definitions for hierarchical classification.}
    \label{tab:definitionsA}
\end{table}

\begin{figure}
    \centering
    \includegraphics[width=\linewidth]{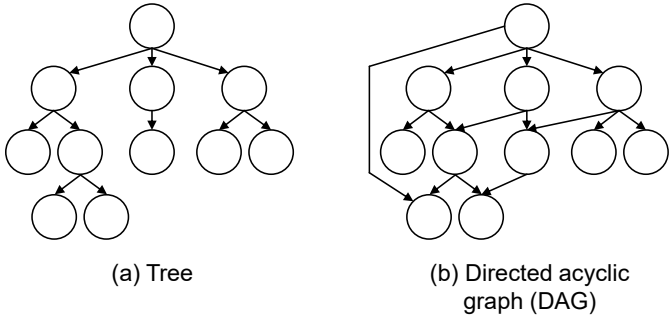}
    \caption{\textbf{Structures of hierarchical classification.}    (a) trees. (b) directed acyclic graphs (DAG). }
    \label{fig:structures}
\end{figure}

% Sub problems of hierarchical classification
According to~\cite{silla2011survey,borges2013evaluation,kosmopoulos2015evaluation}, the different sub problems of hierarchical classification can be categorized based on three aspects: structure $\Gamma$, number of paths $\Psi$, and label depth $\Phi$. 
First, the literature commonly differentiates two types of structures $\Gamma$ that put the classes into a hierarchical relationship: trees $T$ and directed acyclic graphs $DAG$~\cite{silla2011survey}. Both of these structures are  presented in Fig.~\ref{fig:structures}. A tree is a graph in which any two nodes are connected by exactly one path, whereas a DAG allows nodes to possess multiple parental nodes. T implies $\left | W_d \right | = 1$ and DAG entails $\left | W_d \right | \geq 1$ for a single class resp. single labelled object $o_d$. Moreover, this work assumes that a classifier is required to predict a specific path for DAG problems, while T problems require the prediction of a class only.
Second,~\cite{silla2011survey} distinguishes single (SPL) and multi path label (MPL) hierarchical classification depending on the number of allocated paths $\Psi = \left | P_d \right |$ per object $o_d \in \Omega$. SPL implies $\left | P_d \right | = 1$, while MPL entails $\left | P_d \right | \geq 1$.
Third, hierarchical classification problems differ by their required label depth $\Phi$; there are problems that necessarily require full depth labeling (FD) meaning the mandatory leaf-node prediction (MLNP), and such that do not (partial depth labeling, PD or non-mandatory leaf-node prediction, NMLNP)~\cite{freitas2007tutorial,costa2007review}. 

\begin{figure}
    \centering
    \includegraphics[width=\linewidth]{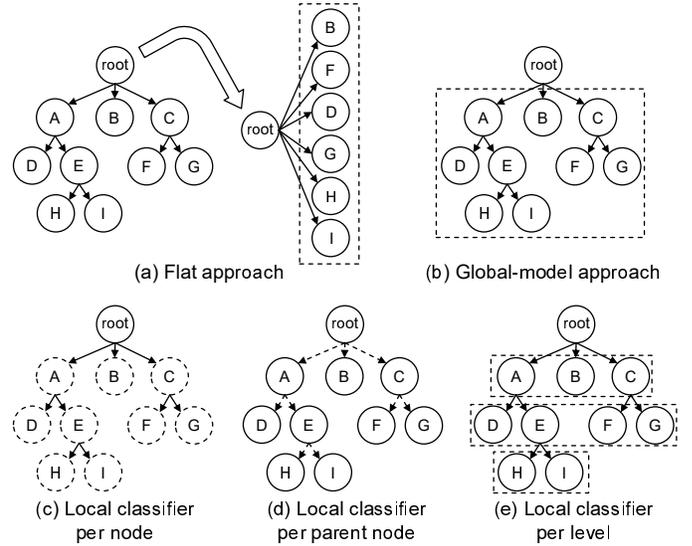}
    \caption{\textbf{Approaches to hierarchical classification.}    The use of classifiers is marked with dashed lines. (a) transforms leaf node classification to a multi-class problem. (b) considers all classes at the same time. (c) employs a binary classifier for each vertex of the structure. (d) utilizes a multi-class classifier for each parental node. (e) uses a multi-class classifier at each level of the taxonomy.}
    \label{fig:hierarchicalApproaches}
\end{figure}

% Approaches / algorithms to hierarchical classification
Algorithms approaching the hierarchical classification problem can be grouped into three categories~\cite{silla2011survey,dumais2000hierarchical, kiritchenko2006learning}: flat, local-model, and global-model approaches. 
These three approaches and their sub types are outlined in Fig.~\ref{fig:hierarchicalApproaches}. 

The flat approach transforms the hierarchical classification problem into a multi-class problem using the leaf-nodes only.
Three types of local-model approaches are mentioned by~\cite{silla2011survey}: local classifier per node, local classifier per parent node, and local classifier per level. 
The local classifier per node approach employs a binary classifier for each node of the class hierarchy, while the local classifier per parent approach trains a multi-class classifier for all non-leaf nodes. 
The local classifier per level approach consists in using a multi-class classifier for all labels present in the levels of a hierarchical structure.
The global approach utilizes one single classifier that captures the whole hierarchical class structure.
\cite{kiritchenko2006learning} argues, that local-model approaches produce consistent labeling and are efficient to train. This comes at the cost that local models are highly sensitive to ancestral decisions and hardly interpretable due to the vast amount of different classifiers. Therefore, they pronounce for the use of global models that probably have a higher chance for correct predictions. 
However, the training of global-model approaches is complex and contrary to the other approaches the classification problem is not decomposed to simpler problems. 
In general, the performance of classifiers depends on their training step. A larger amount of data available and less contradictions in the data set supports the training of better models. This is an argument for local classifiers per level and flat approaches. Local classifiers per node and parent node face a shrinking size of training data for the deeper levels, while global-model approaches face contradicting data as there is no decomposition of the data for different classifiers.

All approaches have their point and with regards to the diversity of algorithms found in the literature, one can conclude that different types of data and hierarchical classification problems require different, suitable approaches.

%%%%%%%%%%%%%%%%%%%%%%%%%%%%%%%%%%%%%%%%%%%%%%%%%%%%%%%%%%%%%%%%%%%%%%%%%%%%%%%%%%%%%%
%%%%%%%%%%%%%%%%%%%%%%%%%%%%%%%%%%%%%%%%%%%%%%%%%%%%%%%%%%%%%%%%%%%%%%%%%%%%%%%%%%%%%%
%%%%%%%%%%%%%%%%%%%%%%%%%%%%% Related Works %%%%%%%%%%%%%%%%%%%%%%%%%%%%%%%%%%%%%%%%%%
%%%%%%%%%%%%%%%%%%%%%%%%%%%%%%%%%%%%%%%%%%%%%%%%%%%%%%%%%%%%%%%%%%%%%%%%%%%%%%%%%%%%%%
%%%%%%%%%%%%%%%%%%%%%%%%%%%%%%%%%%%%%%%%%%%%%%%%%%%%%%%%%%%%%%%%%%%%%%%%%%%%%%%%%%%%%%
\section{Related Works} \label{sec:relatedwork}

% Binary classification measures
In the context of the binary classification problem, there are four possible cases that occur: a true positive prediction (TP), a true negative prediction (TN), a false positive prediction (FP), and a false negative prediction (FN). 
Based on the number of times these four cases occur, that make up the confusion matrix, many different performance measures have been proposed~\cite{fawcett2006introduction}.
Amongst these measures there are accuracy (ACC), precision (PPV), recall (TPR), false negative rate (FNR), false positive rate (FPR), specificity (TNR), prevalence (PT), F1-score (F1), and Matthews correlation coefficient (MCC) to name a few; their definitions are listed in Table~\ref{tab:binaryMeasures}.
\begin{table} [h!]
    \centering
    % \begin{tabular}{l l l}
    %     \textbf{Measure} & \textbf{Symbol} & \textbf{Definition} \\ \hline
    %     Accuracy & $ACC$ & $= \frac{TP+TN}{TP+TN+FP+FN}$ \\
    %     Precision & PPV & $= \frac{TP}{TP+FP}$ \\
    %     Recall & TPR & $= \frac{TP}{TP+FN}$ \\
    %     False neg. rate & FNR & $= \frac{FP}{FN+TP}$ \\
    %     False pos. rate & FPR & $= \frac{FP}{FP+TN}$ \\
    %     Specificity & TNR & $= \frac{TN}{TN+FP}$ \\
    %     Prevalence & PT & $= \frac{\sqrt{TPR(1-TNR)}+TNR-1}{TPR+TNR-1}$ \\
    %     F1-score & F1 & $= \frac{2TP}{2TP+FP+FN}$ \\
    %     Matthews corr. & MCC & $= \frac{TP \times TN - FP \times FN}{\sqrt{(TP+FP)(TP+FN)(TN+FP)(TN+FN)} }$
    \begin{tabular}{l l}
        \textbf{Measure} & \textbf{Definition} \\ \hline
        ACC & $= \frac{TP+TN}{TP+TN+FP+FN}$ \\
        PPV & $= \frac{TP}{TP+FP}$ \\
        TPR & $= \frac{TP}{TP+FN}$ \\
        FNR & $= \frac{FP}{FN+TP}$ \\
        FPR & $= \frac{FP}{FP+TN}$ \\
        TNR & $= \frac{TN}{TN+FP}$ \\
        PT  & $= \frac{\sqrt{TPR(1-TNR)}+TNR-1}{TPR+TNR-1}$ \\
        F1  & $= \frac{2TP}{2TP+FP+FN}$ \\
        MCC & $= \frac{TP \times TN - FP \times FN}{\sqrt{(TP+FP)(TP+FN)(TN+FP)(TN+FN)} }$
        \\ \hline
    \end{tabular}
    \caption{Common binary classification measures.}
    \label{tab:binaryMeasures}
\end{table}

% Requirements to hierarchical classification measures
As it turns out, hierarchical classification is a more complex problem, and the definition of the confusion matrix elements is not trivial. \cite{kiritchenko2006learning} pleads for the consideration of partially correct classification, the discriminatory punishment of more distant errors and errors at upper levels. Take the snack product categorization in Fig.~\ref{fig:classificationProblems}(d) as an example; the wrong classification of \textit{white chocolate} could be considered to be more accurate compared to the wrong classification of \textit{muffins} or even a snack from different flavors.

\cite{kosmopoulos2015evaluation} discusses further issues, namely over- and under-specialization, alternative paths, pairing, and long distance problems. 
With regards to our snack product category example, over- and under-specialization could occur in NMLNP problems, where the classifier predicts \textit{dark chocolate}, and the true label is set as \textit{sweet} snack or the over way around.
Moreover, the MPL classification problem can increase the complexity of evaluation further.

% Hierarchical classification measures
Due to these complexities, specific measures for the evaluation of hierarchical classification problems have been proposed. 
\cite{kosmopoulos2015evaluation} differentiates pair based and set based measures.
Pair based measures assign costs to pairs of predicted and true classes; these measures are most suitable for tree and SPL problems. A possible extension to MPL problems is the relaxation for each class to be part of more than one pair. 
Set based measures take the whole set of predicted and true classes into account, including ancestry and descendance relationships. Symmetric difference loss and hierarchical precision and recall~\cite{kiritchenko2005functional} count amongst the set based measures.

\cite{costa2007review} categorizes the evaluation measures into distance based, depth based, semantics based, and hierarchy based measures.
Distance based measures consider the distance of true and predicted classes in a structure~\cite{wang1999building,sun2001hierarchical}. A possible issue with these measures is the neglection of different levels, as these measures commonly do not consider the difficulty for accurate classification at deeper levels and the severity of misclassification at shallower levels.

A possible mean to encounter this shortcoming is the weighting of structural edges which is described as depth based measures. Besides the number of edges, and the depth of true and predicted classes are taken into consideration. However, the new problem of setting the weights occurs. \cite{blockeel2002hierarchical,holden2006hierarchical} employ exponential weighting. A possible issue with these measures is the discrimination and over-penalization of leaf nodes with different depths. In addition to that, the definition of depth becomes more challenging in DAGs.

Semantics based measures employ the concept of similarity between classes and objects to predict the performance of classification models~\cite{sun2001hierarchical}. From this perspective, similarity or distance can be quantified comparing different feature vectors in a multidimensional space similar to the hierarchical clustering problems~\cite{gauch1981hierarchical,gordon1987review}. One drawback doing so is that during the comparison of two algorithms with different features but identical predictions, evaluation results in different performances.
Hierarchy based measures consider structure related information. Mainly, these measures consider the intersection of allocation paths, with a focus eiter on descendants~\cite{ipeirotis2001probe} or ancestors~\cite{kiritchenko2004hierarchical}.

Finally, we also find loss function based evaluation measures in the literature~\cite{cesa2006hierarchical,cesa2005incremental}, that employ functions similar to the model training for evaluation purposes.

%%%%%%%%%%%%%%%%%%%%%%%%%%%%%%%%%%%%%%%%%%%%%%%%%%%%%%%%%%%%%%%%%%%%%%%%%%%%%%%%%%%%%%
%%%%%%%%%%%%%%%%%%%%%%%%%%%%%%%%%%%%%%%%%%%%%%%%%%%%%%%%%%%%%%%%%%%%%%%%%%%%%%%%%%%%%%
%%%%%%%%%%%%%%%%%%%%%%%%%%%%% Hierarchical Confusion Matrix %%%%%%%%%%%%%%%%%%%%%%%%%%
%%%%%%%%%%%%%%%%%%%%%%%%%%%%%%%%%%%%%%%%%%%%%%%%%%%%%%%%%%%%%%%%%%%%%%%%%%%%%%%%%%%%%%
%%%%%%%%%%%%%%%%%%%%%%%%%%%%%%%%%%%%%%%%%%%%%%%%%%%%%%%%%%%%%%%%%%%%%%%%%%%%%%%%%%%%%%
\section{Hierarchical Confusion Matrix} \label{sec:hConfusion}

% Orientation on confusion matrix
The orientation on binary classification evaluation measures in the context of hierarchical classification problems can be observed in many works such as~\cite{sun2001hierarchical,kiritchenko2005functional,kiritchenko2006learning,dumais2000hierarchical,sokolova2009systematic}.
Especially the measures hierarchical precision (hP), recall (hR), and F-score (hF)~\cite{kiritchenko2005functional,kiritchenko2006learning} enjoy increased attention in the ongoing discussion on hierarchical classification evaluation and are frequently used in studies, e.g.~\cite{borges2013evaluation}. Their definition for SPL problems is shown below, including a degree of freedom meaning that hP is considered $\beta$ times more important than hR:

\begin{flalign}
    \begin{aligned} \label{eq:hPhRhF}
        & hP = \frac{\sum_i \left | \widetilde{P}_i \cap \hat{P}_i \right | }{\sum_i \left | \hat{P}_i \right |} 
        & \quad hR = \frac{\sum_i \left | \widetilde{P}_i \cap \hat{P}_i \right | }{\sum_i \left | \widetilde{P}_i \right |} &
        \\ & hF = \frac{ (1 + \beta^2) \cdot hP \cdot hR }{ (\beta^2 \cdot hP + hR) } &
    \end{aligned}
\end{flalign}

\cite{silla2011survey} recommends the usage of these measures, as the applicability and possible extension to both, trees \& DAGs, and to both, SPL \& MPL problems, is an argument for their choice.
Unmistakably, they resemble strongly to the binary classification measures PPV, TPR, and F1 based on the confusion matrix:

\begin{flalign}
    \begin{aligned} \label{eq:PRF_flatForms}
        & PPV = \frac{TP}{TP+FP} &
        \quad TPR = \frac{TP}{TP+FN} & 
        \\ & F1 = \frac{(1 + \beta^2) \cdot PPV \cdot TPR}{(\beta^2 \cdot PPV + TPR)} &
    \end{aligned} 
\end{flalign}

% Motivation
Open questions arise on the definition of the confusion matrix for hierarchical classification problems. The observation for hP, hR, and hF motivates to propose a novel confusion matrix for hierarchical classification, that opens the door for the application of binary classification measures while considering the peculiarities of the hierarchical classification problem. 
The concept of the hierarchical confusion matrix defines the four fields of the confusion matrix with a connotation to decisions that are done along the prediction path(s) as follows:
\begin{itemize}
  \item \textbf{True positives ($TP_H$)} - the number of nodes that were correctly labeled positively by the predicted path. %= span of the longest common path
  \item \textbf{True negatives ($TN_H$)} - the number of nodes that were correctly labeled negatively by the predicted path. %= neighbors of nodes on the common path and its leaf nodes truly negative predicted descendants 
  \item \textbf{False positives ($FP_H$)} - the number of nodes that were wrongly labeled positively by the predicted path. %= the span of the predicted path minus $TP_H$
  \item \textbf{False negatives ($FN_H$)} - the number of nodes that were wrongly labeled negatively by the predicted path. %= the span of the true path minus $TP_H$
\end{itemize}

After determining $TP_H$, $TN_H$, $FP_H$, and $FN_H$ for a single $o_d \in \Omega$ as a function of $P_d$ and $W_d$, the values of hierarchical confusion can be calculated as the sum over the whole data set and used to determine the ``flat'' evaluation measures of binary classification (see Table~\ref{tab:binaryMeasures}):
\begin{flalign}
    \begin{aligned} \label{eq:PRF_flatForms}
        & TP = \sum_{o_d \in \Omega} TP_H \left \{ P_d, W_d \right \} \quad &
        & TN = \sum_{o_d \in \Omega} TN_H \left \{ P_d, W_d \right \} & \\
        & FP = \sum_{o_d \in \Omega} FP_H \left \{ P_d, W_d \right \} & \quad
        & FN = \sum_{o_d \in \Omega} FN_H \left \{ P_d, W_d \right \} &
    \end{aligned} 
\end{flalign}

In the following we formalize this concept of hierarchical confusion beginning from its simplest form for ($\Gamma$=T, $\Psi$=SPL, $\Phi$=MLNP) problems and then extend it to a generalized form that is applicable to all hierarchical classification problems. 
        
\subsection{Hierarchical Confusion for ($\Gamma$=T, $\Psi$=SPL, $\Phi$=MLNP)}
According to the verbal definition of the confusion matrix elements in the prior section, we formalize the elements as follows.
Since ($\Gamma$=T, $\Psi$=SPL, $\Phi$=MLNP) problems imply $W_d = \widetilde{P}_d$ and $P_d = \hat{P}_d$, we define the elements as function of a given $\widetilde{P}_d$ and $\hat{P}_d$ instead of a given $W_d$ and $P_d$.
The true positive ($TP_H$) equals the number of nodes in the intersection of $\hat{P}_d$ and $\widetilde{P}_d$.
The true negative ($TN_H$) is given by the union of the neighbors of each node on the common path\footnote{The union of the neighbors of each node on the common path excluding nodes of the true path or duplicates in this set. In more complex problems such as DAG it is possible that nodes occur multiple times  in the union of the neighbors of each node on the common path, therefore they are excluded. Moreover, it is possible that nodes of the true path appear in this union set, and therefore are explicitly removed from the set.} and the number of (correctly) negative predicted descendants of the common path's leaf node. 
The number of (correctly) negative predicted descendants of the common path's leaf node can be obtained by the set of descendants of the common path's leaf node excluding the false negative node of the true path $\widetilde{P}_d$ and the false positive node of the predicted path $\hat{P}_d$.

The false positive ($FP_H$) equals the number of nodes of the predicted path that are not part of the common path.
The false negative ($FN_H$) is given by the number of nodes of the true path that are not part of the common path. An example for the evaluation of a prediction in the context of a hierarchical tree classification problem is outlined in Fig.~\ref{fig:hierconfA}(a). To summarize, the four aforementioned cases are formalized below:

\begin{flalign}
    \begin{aligned} \label{eq:PRF_flatForms}
        & TP_H \left \{ \widetilde{P}_d , \hat{P}_d \right \} & = & \left | \hat{P}_d \cap \widetilde{P}_d \setminus R \right | & \\
        & TN_H \left \{  \widetilde{P}_d , \hat{P}_d  \right \} & = & \left | \left \{ \bigcup_{C_i \in \hat{P}_d \overline{\cap} \widetilde{P}_d} N\left \{ C_i \right \} \right \} \setminus \widetilde{P}_d \right | &\\ & & & +  \left | D\left \{ Z\left \{ \hat{P}_d \: \overline{\cap} \: \widetilde{P}_d \right \} \right \} \setminus \left \{ \widetilde{P}_d \cup \hat{P}_d \right \} \right | & \\
        & FP_H \left \{  \widetilde{P}_d , \hat{P}_d  \right \} & = & \left | \hat{P}_d \setminus \widetilde{P}_d \right | & \\
        & FN_H \left \{  \widetilde{P}_d , \hat{P}_d  \right \} & = & \left | \widetilde{P}_d \setminus \hat{P}_d \right | & \\
        %& FP_H \left \{ o_d \right \} & = & \left | \hat{P}_d \right | - \left | \hat{P}_d \cap \widetilde{P}_d \right | & \\
        %& FN_H \left \{ o_d \right \} & = & \left | \widetilde{P}_d \right | - \left | \hat{P}_d \cap \widetilde{P}_d \right | & \\
    \end{aligned}
\end{flalign}

\subsection{Hierarchical Confusion for ($\Psi$=SPL, $\Phi$=MLNP)}
In order to extend any definition for $TP_H$, $TN_H$, $FP_H$, and $FN_H$ in the context of a T to a DAG problem, we propose a two step ``benevolent'' approach. 

\begin{itemize}
  \item \textbf{Step 1}: select the path $i$ ($\widetilde{P}_i \in W_d$) that resembles most to $\hat{P}_d$ (SPL problems require $\left | P_d \right | = 1$): 
\end{itemize}
\begin{equation}
    i = \displaystyle \max_{\substack{ \widetilde{P}_i \in W_d}} \left | \widetilde{P}_i \: \overline{\cap} \: \hat{P}_d \right |
\end{equation}

\begin{itemize}
    \item \textbf{Step 2}: calculate $TP_H$, $TN_H$, $FP_H$, and $FN_H$ like for a ($\Gamma$=T, $\Psi$=SPL, $\Phi$=MLNP) problem using $\widetilde{P}_i$ and $\hat{P}_d$.
\end{itemize}

\begin{figure}
    \centering
    \includegraphics[width=\linewidth]{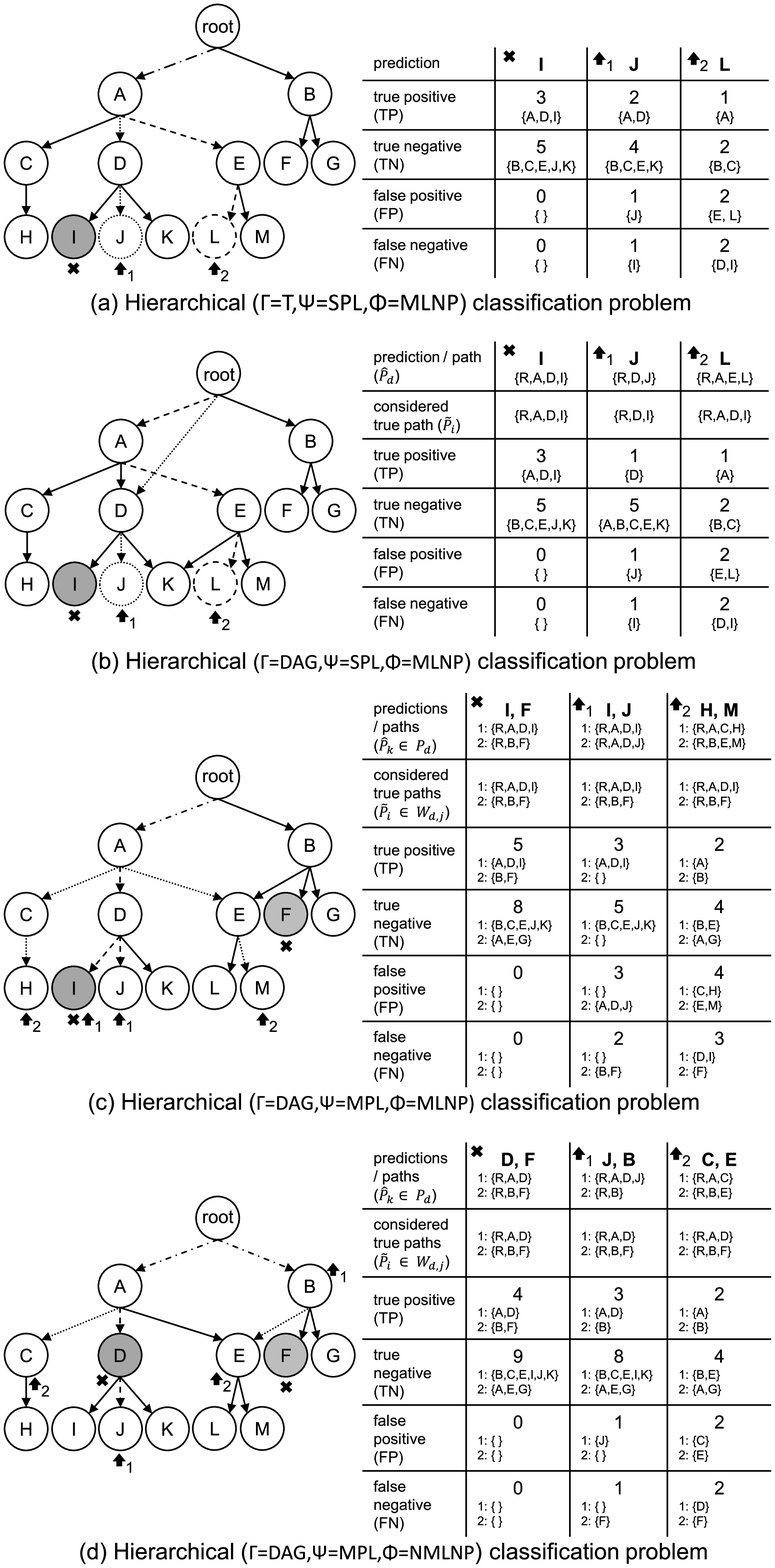}
    \caption{\textbf{Four exemplary predictions and their evaluation using the concept of hierarchical confusion.} The true class(es) is (are) coloured gray and marked with a cross. The predictions are marked with numbered arrows and prediction paths are marked with an dotted resp. dashed lines. The tables right to the diagrams present the resulting numbers for the confusion matrix fields and the relevant nodes that were considered in the specific context. }
    \label{fig:hierconfA}
\end{figure}

We call this approach ``benevolent'' as one $\widetilde{P}_i$ of the possible true paths in $W_d$ is chosen to enable the achievement of the best possible performance for a given prediction $\hat{P}_d$. An example for the evaluation of a ($\Gamma$=DAG) problem is outlined in Fig~\ref{fig:hierconfA}(b).

\subsection{Hierarchical Confusion for ($\Phi$=MLNP)}
The next step towards a generalized hierarchical confusion concept is the extension to MPL problems. Similar to the DAG extension, we can decompose the MPL problem to SPL problems. 
To do so, we first need to extend the notation. 
In MPL problems, $C\left \{ o_d \right \}$ is a set of $n$ true classes of $o_d$ in $\Gamma$. 
The classifier predicts $m$ paths. The number of predicted paths $m$ and the number of true classes $n$ are not necessarily equal in applications, as $n$ is considered to be unknown to the classifier a priori. 
$W_{d,j}$ is the set of all true allocation paths for $o_d$ connected to $C_j \in C\left \{ o_d \right \}$. 
Tree problems imply $\left | W_{d,j}  \right | = 1$. 
Based on this redefinition, the concept of confusion matrix can be extended to MPL problems with following, ``benevolent'' approach:

\begin{itemize}
    \item \textbf{Step 1}: Create a map $M$, that stores values $m = M\left [ \hat{P}_k \right ]$ for each $\hat{P}_k \in P_d$. 
    \item \textbf{Step 2}: For each $\hat{P}_k \in P_d$ determine the $m$ values as follows:
    \begin{itemize}
        \item \textbf{Step 1.1}: Find the path $i$ of any class $j$ in $W_{d,j}$ that resembles most to $\hat{P}_k$:
    \end{itemize}
\begin{equation}
    (i,j) = \displaystyle \max_{\substack{ \widetilde{P}_i \in W_{d,j}}} \left | \widetilde{P}_i \: \overline{\cap} \: \hat{P}_k \right |
\end{equation}
\begin{itemize}
        \item \textbf{Step 1.2}: Set $m$ accordingly to the number of intersections between $\hat{P}_k$ and the found $\widetilde{P}_i \in W_{d,j}$:
    \end{itemize}
\end{itemize}
\begin{equation}
    m = \left | \widetilde{P}_i \: \overline{\cap} \: \hat{P}_k \right | \Rightarrow M\left [ \hat{P}_k \right ]
\end{equation}
\begin{itemize}
    \item \textbf{Step 3}: For each $\hat{P}_k \in P_d$ (in descending order of related $m$ values stored in $M$) determine $TP_{H,k}$, $TN_{H,k}$, $FP_{H,k}$, and $FN_{H,k}$ as follows:
    \begin{itemize}
        \item \textbf{Step 3.1}: If $W_{d,j} \neq \varnothing$, go to Step 3.2. Otherwise, if $W_{d,j}$ is empty, if there are more predictions $m$ than true classes $n$, consider the whole prediction as a false positive:
    \end{itemize}
\begin{flalign}
    \begin{aligned} \label{eq:PRF_flatForms}
        & TP_{H,k} = 0 \quad &
        & TN_{H,k} = 0 & \\
        & FP_{H,k} = \left | \hat{P}_k \setminus R \right | & \quad
        & FN_{H,k} = 0 &
    \end{aligned} 
\end{flalign}
\begin{itemize}
        \item \textbf{Step 3.2}: Find the path $i$ of any class $j$ in $W_{d,j}$ that resembles most to $\hat{P}_k$:
    \end{itemize}
\begin{equation}
    (i,j) = \displaystyle \max_{\substack{ \widetilde{P}_i \in W_{d,j}}} \left | \widetilde{P}_i \: \overline{\cap} \: \hat{P}_k \right |
\end{equation}
\begin{itemize}
        \item \textbf{Step 3.3}: Calculate $TP_{H,k}$, $TN_{H,k}$, $FP_{H,k}$, and $FN_{H,k}$ like a ($\Gamma$=T, $\Psi$=SPL, $\Phi$=MLNP) problem using $\widetilde{P}_i$ and $\hat{P}_k$.

        \item \textbf{Step 3.4}: Remove all paths in $W_{d,j}$ related to class $j$. Remove $C_j$ from the set of true labels $C\left \{ o_d \right \}$. 
\end{itemize}
\end{itemize}
\begin{itemize}
    \item \textbf{Step 4}: Calculate the fields of the confusion matrix as sum of values determined by the single predictions $\hat{P}_k \in P_d$. For all remaining classes $C_j$ that were not removed from $C\left \{ o_d \right \}$ (if there are more true classes $n$ than predictions $m$), consider the shortest paths inside as false negatives:
\end{itemize}
\begin{flalign}
    \begin{aligned} \label{eq:PRF_flatForms}
        & TP_H = \sum_{\hat{P}_k \in P_d} TP_{H,k} \quad
        & TN_H = \sum_{\hat{P}_k \in P_d} TN_{H,k} \\
        & FP_H = \sum_{\hat{P}_k \in P_d} FP_{H,k} & \\
        & FN_H = \sum_{\hat{P}_k \in P_d} FN_{H,k} & + \sum_{C_j \in C\left \{ o_d \right \}}  \min_{ \forall \widetilde{P}_i \in W_{d,j}} \left | \widetilde{P}_i \setminus R \right |
    \end{aligned} 
\end{flalign}

An example for the evaluation of a ($\Gamma$=DAG, $\Psi$=MPL) problem is depicted in Fig.~\ref{fig:hierconfA}(c).

\subsection{Generalized Hierarchical Confusion (for all problems)}
Finally, the discussion of NMLNP problems remains. As it turns out, the approach presented for MPL problems is applicable to NMLNP problems as well, as shown in Fig.~\ref{fig:hierconfA}(d). NMLNP adds complexity to the game by not requiring the prediction to the deepest level of the taxonomy. This allows for misclassification errors such as over- and under-specialization. The discussed approach in the previous subsection is also suitable in this case, as the confusion matrix element definitions consider paths and their intersection.

%%%%%%%%%%%%%%%%%%%%%%%%%%%%%%%%%%%%%%%%%%%%%%%%%%%%%%%%%%%%%%%%%%%%%%%%%%%%%%%%%%%%%%
%%%%%%%%%%%%%%%%%%%%%%%%%%%%%%%%%%%%%%%%%%%%%%%%%%%%%%%%%%%%%%%%%%%%%%%%%%%%%%%%%%%%%%
%%%%%%%%%%%%%%%%%%%%%%%%%%%%% Experiments %%%%%%%%%%%%%%%%%%%%%%%%%%%%%%%%%%%%%%%%%%%%
%%%%%%%%%%%%%%%%%%%%%%%%%%%%%%%%%%%%%%%%%%%%%%%%%%%%%%%%%%%%%%%%%%%%%%%%%%%%%%%%%%%%%%
%%%%%%%%%%%%%%%%%%%%%%%%%%%%%%%%%%%%%%%%%%%%%%%%%%%%%%%%%%%%%%%%%%%%%%%%%%%%%%%%%%%%%%
\section{Experiments} \label{sec:experiment}
In this section, we apply our concept of the hierarchical confusion matrix onto three, real-world hierarchical classification problems with differing levels of complexity. Employing binary performance measures (see Table~\ref{tab:binaryMeasures}) using the hierarchical confusion matrix allows for the comparison of different classification models in the three experiments. Code and data for reproduction can be found on GitHub.

\subsection{Transposon Classification}
Transposon classification is a task from bioinformatics and computational biology~\cite{riehl2021}. Transposons are DNA sequences that are able to move inside the genome using copy \& paste and cut \& paste transposition mechanisms and are suspected to play an important role as driver for genome evolution. Moreover, they can be classified into a tree structure depending on structural and protein features obtained by the sequences that determine their specific transposition mechanisms.
The transposon classification task is a ($\Gamma$=T, $\Psi$=SPL, $\Phi$=MLNP) classification problem.
Within this experiment we use the combined dataset mipsREdat-PGSB\footnote{\url{https://pgsb.helmholtz-muenchen.de/plant/recat/index.jsp}}, RepBase v23.08\footnote{\url{https://www.girinst.org/repbase/}}, and the taxonomy proposed in~\cite{riehl2021}.

Table~\ref{tab:experiment_transposons} presents the hierarchical confusion matrix and evaluation measures based on it for different classification models that were compared in~\cite{riehl2021}. Situations in which a clear ranking based on one measure is complicated, the consideration of multiple measures as presented in the table can facilitate model ranking. The results show that evaluation measures based on the novel hierarchical confusion matrix span the whole range, comparable to the related binary measures.

\subsection{GermEval 2019 competition, Task 1A}
The GermEval 2019 competition~\cite{remus2019germeval} was a competition hold by the Language Technology Group of Universität Hamburg, Germany. In total 18 teams participated to the competition and worked on multiple tasks. Here, we consider Task 1A and 1B, related to hierarchical classification of blurb texts in German language - short descriptions of books. The data set covers 20,784 books and a 3-level DAG structure with 343 categories. All participant submissions, data set and taxonomy were downloaded from the competition's website\footnote{\url{https://www.inf.uni-hamburg.de/en/inst/ab/lt/resources/data/germeval-2019-hmc/germeval2019t1-public-data-final.zip}}.

Task 1A is a ($\Gamma$=(1-level)T, $\Psi$=MPL, $\Phi$=MLNP) classification problem. The challenge excercise was to classify one or multiple top-level categories for the blurbs that represent literature genres.  
Table~\ref{tab:experiment_germeval1a} summarizes the confusion matrix, evaluation measures calculated by us (HC) and the reported performances by the organizers of the competition (GE) for each participant and two baseline models (Baseline, Contender). The organizers employed micro-averaged recall, precision, and F1-score~\cite{silla2011survey, sorower2010literature} as well as subset accuracy~\cite{sorower2010literature} for evaluation purposes. This table allows for the comparison of the reported evaluation measures. Also, it enables the comparison of different rankings based on the reported evaluation measures. Additionally, the MCC score based on the hierarchical confusion matrix is reported.

First of all, the evaluation results and rankings for both evaluation approaches resemble strongly. The rankings differ by only one pair (Comtravo-DS and HSHL). 
Next, the differences in precision, recall, and F1 score between the hierarchical confusion matrix and micro-averaged measures are relatively small (on average  2.41\%, 0.10\% and 1.44\%). However, larger differences become obvious with regards to accuracy (on average 17.53\%). Differences in accuracy could be explained by the different perception of true negatives in our proposed confusion matrix and subset accuracy.

\subsection{GermEval 2019 competition, Task 1B}
Task 1B is a ($\Gamma$=DAG, $\Psi$=MPL, $\Phi$=NMLNP) classification problem. The challenge excercise was to classify one or multiple categories from the DAG for the blurbs, not necessarily to the deepest level. 
Table~\ref{tab:experiment_germeval1b} summarizes the results similar to the task before. Again, resulting rankings are identical except for one pair (EricssonResearch and TwistBytes). Differences in accuracy, precision, recall, and F1 score (67.44\%, 2.44\%, 5.20\%, 4.25\%) are higher compared to the previous task. Again, accuracy differs most. As Task 1B considers a more complex structure, differences between the evaluation measures become more obvious.

\begin{table*}[t]
  \centering
\begin{tabular}{l|rrrr|rrrrr|c}
\textbf{model} & \textbf{TP} & \textbf{TN} & \textbf{FP} & \textbf{FN} & \textbf{ACC} & \textbf{PPV} & \textbf{TPR} & \textbf{F1} & \textbf{MCC} & \textbf{Rank} \\
\hline
HC\_GA        & 19,145      & 27,690      & 7,854       & 7,744       & 75.02\%      & 70.91\%      & 71.20\%      & 71.05\%     & 49.08\%      & 2             \\
HC\_LGA       & 18,420      & 25,582      & 8,598       & 8,469       & 72.05\%      & 68.18\%      & 68.50\%      & 68.34\%     & 43.33\%      & 3             \\
NLLCPN        & 17,608      & 24,765      & 9,410       & 9,281       & 69.39\%      & 65.17\%      & 65.48\%      & 65.33\%     & 37.93\%      & 4             \\
RFSB          & 22,833      & 34,026      & 4,131       & 4,056       & 87.41\%      & 84.68\%      & 84.92\%      & 84.80\%     & 74.06\%      & 1             \\
TERL          & 366         & 630         & 18,776      & 26,523      & 2.15\%       & 1.91\%       & 1.36\%       & 1.59\%      & -95.58\%     & 6             \\
TopDown       & 1,415       & 2,743       & 16,603      & 25,474      & 8.99\%       & 7.85\%       & 5.26\%       & 6.30\%      & -81.49\%     & 5   
\\ \hline
\end{tabular}
  \caption{Transposon Classification Experiment.}
  \label{tab:experiment_transposons}
\end{table*}

\begin{table*}[t]
\centering
\resizebox{\textwidth}{!}{
\begin{tabular}{l|rrrr|rrrrr|rrrr|cc}
                 & \multicolumn{9}{c}{\textbf{Hierarchical confusion matrix performance evaluation (HC)}}                                          & \multicolumn{4}{c}{\textbf{Reported performances (GE)}} & \textbf{}          &                    \\
\textbf{model}    & \textbf{TP} & \textbf{TN} & \textbf{FP} & \textbf{FN} & \textbf{ACC} & \textbf{PPV} & \textbf{TPR} & \textbf{F1} & \textbf{MCC} & \textbf{ACC}      & \textbf{PPV}     & \textbf{TPR}     & \textbf{F1}     & \textbf{Rank (HC)} & \textbf{Rank (GE)} \\
\hline
Averbis          & 3,613       & 28,863      & 584         & 857         & 95.75\%      & 86.09\%      & 80.83\%      & 83.37\%     & 80.99\%      & 79.00\%           & 86.00\%          & 81.00\%          & 83.00\%         & 6                  & 6                  \\
Comtravo-DS      & 3,690       & 29,517      & 1,078       & 780         & 94.70\%      & 77.39\%      & 82.55\%      & 79.89\%     & 76.89\%      & 72.00\%           & 81.00\%          & 83.00\%          & 82.00\%         & 7                  & 8                  \\
DFKI-SLT         & 3,787       & 28,933      & 536         & 683         & 96.41\%      & 87.60\%      & 84.72\%      & 86.14\%     & 84.09\%      & 82.00\%           & 88.00\%          & 85.00\%          & 86.00\%         & 3                  & 3                  \\
EricssonResearch & 3,769       & 28,891      & 455         & 701         & 96.58\%      & 89.23\%      & 84.32\%      & 86.70\%     & 84.79\%      & 84.00\%           & 89.00\%          & 84.00\%          & 87.00\%         & 1                  & 1                  \\
Fosil-hsmw       & 3,719       & 29,003      & 694         & 751         & 95.77\%      & 84.27\%      & 83.20\%      & 83.73\%     & 81.30\%      & 79.00\%           & 84.00\%          & 83.00\%          & 84.00\%         & 5                  & 5                  \\
HSHL             & 3,647       & 28,877      & 777         & 823         & 95.31\%      & 82.44\%      & 81.59\%      & 82.01\%     & 79.32\%      & 77.00\%           & 82.00\%          & 82.00\%          & 82.00\%         & 8                  & 7                  \\
HUIU             & 3,608       & 28,808      & 867         & 862         & 94.94\%      & 80.63\%      & 80.72\%      & 80.67\%     & 77.76\%      & 76.00\%           & 81.00\%          & 81.00\%          & 81.00\%         & 9                  & 9                  \\
Raghavan         & 3,747       & 28,983      & 522         & 723         & 96.34\%      & 87.77\%      & 83.83\%      & 85.75\%     & 83.68\%      & 83.00\%           & 88.00\%          & 84.00\%          & 86.00\%         & 4                  & 4                  \\
TwistBytes       & 3,852       & 29,551      & 752         & 618         & 96.06\%      & 83.67\%      & 86.17\%      & 84.90\%     & 82.65\%      & 79.00\%           & 87.00\%          & 86.00\%          & 86.00\%         & 2                  & 2                  \\
\hline
Baseline         & 3,344       & 29,084      & 544         & 1,126       & 95.10\%      & 86.01\%      & 74.81\%      & 80.02\%     & 77.49\%      & 71.00\%           & 86.00\%          & 75.00\%          & 80.00\%         &                    &                    \\
Contender        & 3,809       & 29,835      & 2,301       & 661         & 91.91\%      & 62.34\%      & 85.21\%      & 72.00\%     & 68.53\%      & 74.00\%           & 82.00\%          & 85.00\%          & 84.00\%         &                    &         
\\ \hline
\end{tabular}}
  \caption{GermEval 2019 Competition, Task 1A Experiment.}
  \label{tab:experiment_germeval1a}
\end{table*}

\begin{table*}[t]
\resizebox{\textwidth}{!}{
\begin{tabular}{l|rrrr|rrrrr|rrrr|cc}
                 & \multicolumn{9}{c}{\textbf{Hierarchical confusion matrix performance   evaluation (HC)}}                                        & \multicolumn{4}{c}{\textbf{Reported performances (GE)}} & \textbf{}          &                    \\
\textbf{model}    & \textbf{TP} & \textbf{TN} & \textbf{FP} & \textbf{FN} & \textbf{ACC} & \textbf{PPV} & \textbf{TPR} & \textbf{F1} & \textbf{MCC} & \textbf{ACC} & \textbf{PPV} & \textbf{TPR} & \textbf{F1} & \textbf{Rank (HC)} & \textbf{Rank (GE)} \\
\hline
Averbis          & 8,552       & 125,951     & 4,683       & 6,558       & 92.29\%      & 64.62\%      & 56.60\%      & 60.34\%     & 56.24\%      & 27.00\%      & 68.00\%      & 61.00\%      & 64.00\%     & 3                  & 3                  \\
Comtravo-DS      & 7,187       & 111,871     & 3,376       & 7,923       & 91.33\%      & 68.04\%      & 47.56\%      & 55.99\%     & 52.36\%      & 19.00\%      & 70.00\%      & 53.00\%      & 60.00\%     & 6                  & 6                  \\
DKFI-SLT         & 7,049       & 112,567     & 2,256       & 8,061       & 92.06\%      & 75.75\%      & 46.65\%      & 57.74\%     & 55.56\%      & 21.00\%      & 78.00\%      & 52.00\%      & 62.00\%     & 4                  & 4                  \\
EricssonResearch & 8,498       & 119,546     & 3,208       & 6,612       & 92.88\%      & 72.60\%      & 56.24\%      & 63.38\%     & 60.10\%      & 38.00\%      & 74.00\%      & 62.00\%      & 67.00\%     & 1                  & 2                  \\
HSHL             & 7,167       & 106,488     & 3,025       & 7,943       & 91.20\%      & 70.32\%      & 47.43\%      & 56.65\%     & 53.21\%      & 26.00\%      & 72.00\%      & 54.00\%      & 62.00\%     & 5                  & 5                  \\
TwistBytes       & 9,174       & 130,886     & 4,747       & 5,936       & 92.91\%      & 65.90\%      & 60.71\%      & 63.20\%     & 59.35\%      & 25.00\%      & 71.00\%      & 65.00\%      & 68.00\%     & 2                  & 1                  \\
\hline
Baseline         & 5,183       & 97,128      & 964         & 9,927       & 90.38\%      & 84.32\%      & 34.30\%      & 48.77\%     & 50.00\%      & 15.00\%      & 85.00\%      & 39.00\%      & 53.00\%     &                    &                    \\
Contender        & 7,693       & 118,017     & 2,854       & 7,417       & 92.45\%      & 72.94\%      & 50.91\%      & 59.97\%     & 57.05\%      & 25.00\%      & 76.00\%      & 56.00\%      & 64.00\%     &                    &
\\ \hline
\end{tabular}}
  \caption{GermEval 2019 Competition, Task 1B Experiment.}
  \label{tab:experiment_germeval1b}
\end{table*}

%%%%%%%%%%%%%%%%%%%%%%%%%%%%%%%%%%%%%%%%%%%%%%%%%%%%%%%%%%%%%%%%%%%%%%%%%%%%%%%%%%%%%%
%%%%%%%%%%%%%%%%%%%%%%%%%%%%%%%%%%%%%%%%%%%%%%%%%%%%%%%%%%%%%%%%%%%%%%%%%%%%%%%%%%%%%%
%%%%%%%%%%%%%%%%%%%%%%%%%%%%% Conclusions %%%%%%%%%%%%%%%%%%%%%%%%%%%%%%%%%%%%%%%%%%%%
%%%%%%%%%%%%%%%%%%%%%%%%%%%%%%%%%%%%%%%%%%%%%%%%%%%%%%%%%%%%%%%%%%%%%%%%%%%%%%%%%%%%%%
%%%%%%%%%%%%%%%%%%%%%%%%%%%%%%%%%%%%%%%%%%%%%%%%%%%%%%%%%%%%%%%%%%%%%%%%%%%%%%%%%%%%%%
\section{Conclusions} \label{sec:conclusion}

% Summary what we did
In this work, we proposed the novel concept of hierarchical confusion matrix, 
which has been extended to a generalized form and we proved its applicability to all types of hierarchical classification problems including DAG structures, MPL, and NMLNP problems.
The evaluation concept was demonstrated on real world applications, benchmarking hierarchical classification problems.
The evaluation results were compared to alternative evaluation measures present in literature, revealing and supporting the reasonability of the proposed concept.
The novel concept of hierarchical confusion matrix allows for the evaluation of hierarchical classification problems using measures based on binary classification problems. Thus, it will facilitate future research on evaluation of hierarchical classification models.

%%%%%%%%%%%%%%%%%%%%%%%%%%%%%%%%%%%%%%%%%%%%%%%%%%%%%%%%%%%%%%%%%%%%%%%%%%%%%%%%%%%%%%
%%%%%%%%%%%%%%%%%%%%%%%%%%%%%%%%%%%%%%%%%%%%%%%%%%%%%%%%%%%%%%%%%%%%%%%%%%%%%%%%%%%%%%
%%%%%%%%%%%%%%%%%%%%%%%%%%%%% References %%%%%%%%%%%%%%%%%%%%%%%%%%%%%%%%%%%%%%%%%%%%%
%%%%%%%%%%%%%%%%%%%%%%%%%%%%%%%%%%%%%%%%%%%%%%%%%%%%%%%%%%%%%%%%%%%%%%%%%%%%%%%%%%%%%%
%%%%%%%%%%%%%%%%%%%%%%%%%%%%%%%%%%%%%%%%%%%%%%%%%%%%%%%%%%%%%%%%%%%%%%%%%%%%%%%%%%%%%%
\bibliography{document} 

\begin{thebibliography}{10}

\bibitem{blockeel2002hierarchical}
Hendrik Blockeel, Maurice Bruynooghe, Sa{\v{s}}o D{\v{z}}eroski, Jan Ramon, and
  Jan Struyf.
\newblock Hierarchical multi-classification.
\newblock In {\em Workshop Notes of the KDD'02 Workshop on Multi-Relational
  Data Mining}, pages 21--35, 2002.

\bibitem{borges2013evaluation}
Helyane~Bronoski Borges, Carlos~N Silla~Jr, and J{\'u}lio~Cesar Nievola.
\newblock An evaluation of global-model hierarchical classification algorithms
  for hierarchical classification problems with single path of labels.
\newblock {\em Computers \& Mathematics with Applications}, 66(10):1991--2002,
  2013.

\bibitem{cesa2005incremental}
Nicolo Cesa-Bianchi, Claudio Gentile, Andrea Tironi, and Lucas Zaniboni.
\newblock Incremental algorithms for hierarchical classification.
\newblock {\em Advances in neural information processing systems}, 17:233--240,
  2005.

\bibitem{cesa2006hierarchical}
Nicolo Cesa-Bianchi, Claudio Gentile, and Luca Zaniboni.
\newblock Hierarchical classification: combining bayes with svm.
\newblock In {\em Proceedings of the 23rd international conference on Machine
  learning}, pages 177--184, 2006.

\bibitem{costa2007review}
Eduardo Costa, Ana Lorena, ACPLF Carvalho, and Alex Freitas.
\newblock A review of performance evaluation measures for hierarchical
  classifiers.
\newblock In {\em Evaluation methods for machine learning II: Papers from the
  AAAI-2007 workshop}, pages 1--6, 2007.

\bibitem{dumais2000hierarchical}
Susan Dumais and Hao Chen.
\newblock Hierarchical classification of web content.
\newblock In {\em Proceedings of the 23rd annual international ACM SIGIR
  conference on Research and development in information retrieval}, pages
  256--263, 2000.

\bibitem{fawcett2006introduction}
Tom Fawcett.
\newblock An introduction to roc analysis.
\newblock {\em Pattern recognition letters}, 27(8):861--874, 2006.

\bibitem{freitas2007tutorial}
Alex Freitas and Andr{\'e} Carvalho.
\newblock A tutorial on hierarchical classification with applications in
  bioinformatics.
\newblock {\em Research and trends in data mining technologies and
  applications}, pages 175--208, 2007.

\bibitem{gauch1981hierarchical}
Hugh~G Gauch~Jr and Robert~H Whittaker.
\newblock Hierarchical classification of community data.
\newblock {\em The Journal of Ecology}, pages 537--557, 1981.

\bibitem{gordon1987review}
Allan~D Gordon.
\newblock A review of hierarchical classification.
\newblock {\em Journal of the Royal Statistical Society: Series A (General)},
  150(2):119--137, 1987.

\bibitem{holden2006hierarchical}
Nicholas Holden and Alex~A Freitas.
\newblock Hierarchical classification of g-protein-coupled receptors with a
  pso/aco algorithm.
\newblock In {\em Proceedings of the IEEE Swarm Intelligence Symposium
  (SIS'06)}, pages 77--84. IEEE Press, 2006.

\bibitem{ipeirotis2001probe}
Panagiotis~G Ipeirotis, Luis Gravano, and Mehran Sahami.
\newblock Probe, count, and classify: categorizing hidden web databases.
\newblock In {\em Proceedings of the 2001 ACM SIGMOD international conference
  on Management of data}, pages 67--78, 2001.

\bibitem{kiritchenko2004hierarchical}
Svetlana Kiritchenko, Stan Matwin, A~Fazel Famili, et~al.
\newblock Hierarchical text categorization as a tool of associating genes with
  gene ontology codes.
\newblock In {\em European workshop on data mining and text mining in
  bioinformatics}, pages 30--34, 2004.

\bibitem{kiritchenko2005functional}
Svetlana Kiritchenko, Stan Matwin, A~Fazel Famili, et~al.
\newblock Functional annotation of genes using hierarchical text
  categorization.
\newblock In {\em Proc. of the ACL Workshop on Linking Biological Literature,
  Ontologies and Databases: Mining Biological Semantics}, 2005.

\bibitem{kiritchenko2006learning}
Svetlana Kiritchenko, Stan Matwin, Richard Nock, and A~Fazel Famili.
\newblock Learning and evaluation in the presence of class hierarchies:
  Application to text categorization.
\newblock In {\em Conference of the Canadian Society for Computational Studies
  of Intelligence}, pages 395--406. Springer, 2006.

\bibitem{koller1997hierarchically}
Daphne Koller and Mehran Sahami.
\newblock Hierarchically classifying documents using very few words.
\newblock Technical report, Stanford InfoLab, 1997.

\bibitem{kosmopoulos2015evaluation}
Aris Kosmopoulos, Ioannis Partalas, Eric Gaussier, Georgios Paliouras, and Ion
  Androutsopoulos.
\newblock Evaluation measures for hierarchical classification: a unified view
  and novel approaches.
\newblock {\em Data Mining and Knowledge Discovery}, 29(3):820--865, 2015.

\bibitem{remus2019germeval}
Steffen Remus, Rami Aly, and Chris Biemann.
\newblock Germeval 2019 task 1: Hierarchical classification of blurbs.
\newblock In {\em KONVENS}, 2019.

\bibitem{riehl2021}
Kevin Riehl, Cristian Riccio, Eric Miska, and Martin Hemberg.
\newblock Transposon ultimate: a bundle of tools for transposon identification.
\newblock 2021 (IN PREPARATION, AVAILABLE ON BIORXIV SOON).

\bibitem{silla2011survey}
Carlos~N Silla and Alex~A Freitas.
\newblock A survey of hierarchical classification across different application
  domains.
\newblock {\em Data Mining and Knowledge Discovery}, 22(1):31--72, 2011.

\bibitem{sokolova2006beyond}
Marina Sokolova, Nathalie Japkowicz, and Stan Szpakowicz.
\newblock Beyond accuracy, f-score and roc: a family of discriminant measures
  for performance evaluation.
\newblock In {\em Australasian joint conference on artificial intelligence},
  pages 1015--1021. Springer, 2006.

\bibitem{sokolova2009systematic}
Marina Sokolova and Guy Lapalme.
\newblock A systematic analysis of performance measures for classification
  tasks.
\newblock {\em Information processing \& management}, 45(4):427--437, 2009.

\bibitem{sorower2010literature}
Mohammad~S Sorower.
\newblock A literature survey on algorithms for multi-label learning.
\newblock {\em Oregon State University, Corvallis}, 18:1--25, 2010.

\bibitem{sun2001hierarchical}
Aixin Sun and Ee-Peng Lim.
\newblock Hierarchical text classification and evaluation.
\newblock In {\em Proceedings 2001 IEEE International Conference on Data
  Mining}, pages 521--528. IEEE, 2001.

\bibitem{wang1999building}
Ke~Wang, Senqiang Zhou, and Shiang~Chen Liew.
\newblock Building hierarchical classifiers using class proximity.
\newblock In {\em VLDB}, volume~99, pages 7--10. Citeseer, 1999.

\end{thebibliography}
\bibliographystyle{plain}
\end{document}